\documentclass[letterpaper, 10 pt, conference]{ieeeconf} 

\IEEEoverridecommandlockouts

\usepackage{graphicx} 
\usepackage[utf8]{inputenc}
\usepackage{amsmath,amsfonts,booktabs,cite} 
\usepackage{gensymb} 
\usepackage{siunitx,textcomp} 
\usepackage[hidelinks]{hyperref} 
\usepackage{cleveref} 
\usepackage{subcaption}
\let\labelindent\relax
\usepackage[inline]{enumitem} 

\def\objs{\mathcal{E}}

\def\cob{Care-O-bot 4}
\def\pc{point-cloud}
\def\hmap{Hypermap}
\def\eoffice{\textit{Office}}
\def\elab{\textit{Laboratory}}
\def\efloor{\textit{Floor}}
\def\ie{\textit{i.e.,}}
\def\eg{\textit{e.g.,}}
\def\etal{\textit{et al.}}

\def\vspacefig{\vspace{-.8em}} 

\title{\LARGE \bf%
\hmap{} Mapping Framework\\and its Application to Autonomous Semantic
Exploration}

\author{Tobias~Zaenker, Francesco~Verdoja and Ville~Kyrki%
\thanks{This work was supported by the Strategic Research Council at Academy of
Finland, decision 314180, and partially funded by the Deutsche
Forschungsgemeinschaft (DFG, German Research Foundation) under Germany’s
Excellence Strategy – EXC 2070 – 390732324.} \thanks{T.~Zaenker
(\texttt{tzaenker{@}uni-bonn.de}) is with the Humanoid Robots Lab, University of
Bonn, Germany; F.~Verdoja and V.~Kyrki (\texttt{\{first.surname\}{@}aalto.fi})
are with School of Electrical Engineering, Aalto University, Finland.}}

\begin{document}

\maketitle
\thispagestyle{empty}
\pagestyle{empty}


\begin{abstract}
Modern intelligent and autonomous robotic applications often require robots to
have more information about their environment than that provided by traditional
occupancy grid maps. For example, a robot tasked to perform autonomous semantic
exploration has to label objects in the environment it is traversing while
autonomously navigating. To solve this task the robot needs to at least maintain
an occupancy map of the environment for navigation, an exploration map keeping
track of which areas have already been visited, and a semantic map where
locations and labels of objects in the environment are recorded. As the number
of maps required grows, an application has to know and handle different map
representations, which can be a burden.

We present the \hmap{} framework, which can manage multiple maps of different
types. In this work, we explore the capabilities of the framework to handle
occupancy grid layers and semantic polygonal layers, but the framework can be
extended with new layer types in the future. Additionally, we present an
algorithm to automatically generate semantic layers from RGB-D images. We
demonstrate the utility of the framework using the example of autonomous
exploration for semantic mapping.
\end{abstract}


\section{Introduction}
\label{sec:intro}

In mobile robotics, the most common way to represent spatial information about
the environment is through maps, which differ in precision and complexity
depending on the application. If only navigation capabilities are required, 2D
occupancy grid maps are the most common for indoor robots as they are usually
sufficient for planar navigation. These maps represent the environment as a
fixed size grid where each cell describes the occupancy probability of the area
it represents~\cite{grisetti_improved_2007}. However, as soon as tasks require a
deeper understanding of the environment, occupancy alone is usually not enough
and other properties of the environment need to be recorded in other specific
maps. For example, autonomous robots with task level capabilities often require
qualitative and richer information about the environment maintained in semantic
maps, which assign labels to objects or places in the
environment~\cite{kostavelis_semantic_2015}. More advanced applications may
require reasoning on an even more diverse number and types of
maps~\cite{deeken_grounding_2018, verdoja_deep_2019}. 

An example of these more advanced applications is autonomous semantic
exploration, where a robot's task is to label objects in the environment it is
in while autonomously navigating in it. To solve this task the robot needs at
least to maintain an occupancy map of the environment for navigation, an
exploration map keeping track of which areas have already been visited and a
semantic map where locations and labels of objects in the environment are
recorded. As the number of maps required by an application grows, the
interaction with each map requires to be handled independently as each map
represents the environment using different constructs, which need to be known
and handled by the application.

\begin{figure}
	\centering
	\includegraphics[width=\linewidth]{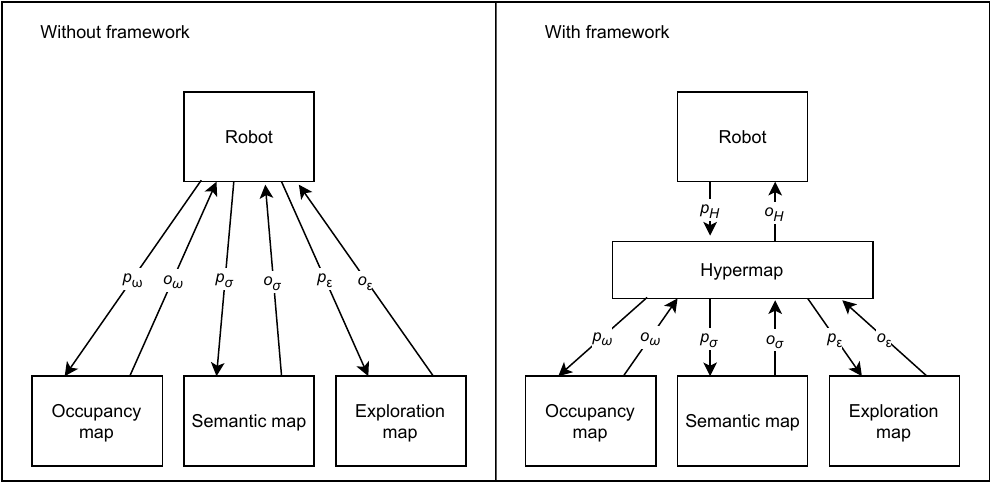}
	\caption{\label{fig:framework_concept}Example of robot-map interaction, with
	and without the \hmap{} framework.}
	\vspacefig{}
\end{figure}

In this work, we propose a framework that integrates and manages multiple maps
of different types. The framework works as an interface that unifies and
simplifies the access to different types of maps for applications.
\Cref{fig:framework_concept} shows the underlying concept. With the framework,
it is not necessary to deal with various representations of different maps.
Instead, unified access to all maps is provided, while the framework handles
conversions internally. This unified access also improves flexibility of the
used maps. For example, a grid based occupancy map could easily be replaced with
a polygonal map, which could be accessed in the same way through the framework.
An application using the hypermap framework would not need to adapt to this
change.

Furthermore, we introduce a mapping algorithm that generates polygonal semantic
maps from RGB-D images for the framework. We also demonstrate that the framework
enables autonomous semantic mapping of arbitrary polygonal areas in an existing
occupancy map.

The main contributions of this work are:
\begin{enumerate*}[label=(\roman*)]
    \item The introduction of a formalism for maps that defines them in terms of
    their functionalities and properties.
	\item A framework able to manage multiple maps of different types, presented
	both theoretically and with a software implementation.
	\item An algorithm to generate polygonal semantic maps from RBG-D images.
	\item A demonstration of the framework's utility by showing its use for
	autonomous semantic exploration of a specified area.
\end{enumerate*}

\section{Related Works}
\label{sec:related}

Occupancy grid maps are the most common type of maps in
robotics~\cite{siegwart_introduction_2011}. These maps, usually generated
automatically through Simultaneous Localization and Mapping (SLAM)
algorithms~\cite{grisetti_improved_2007}, decompose the space into a fixed-size
grid where each cell contains a probability of occupancy of the area
corresponding to that cell. This representation, although convenient, has some
limitations, one of which is that each cell is treated separately and the extent
of an object in the environment cannot intuitively be deduced.

Continuous-valued maps, on the other hand, allow for a more exact representation
of the entities in the environment by representing obstacles as geometric
primitives, \eg{} polygons or lines~\cite{foux_two-dimensional_1993,
zhang_line_2000}. The map complexity is higher than for grid maps, and depends
on the number and shape of obstacles. Often, simplifications are performed to
reduce it, \eg{} by replacing complex polygons with simpler ones or only
displaying features relevant to the sensors and navigation. Several navigation
algorithms \cite{rao1993robot} rely on a polygonal obstacle representation to
find optimal or near-optimal paths. Recent research~\cite{missura_minimal_2018}
also showed interest in polygonal maps due to advantages compared to grid maps,
such as the ability to create visibility graphs for an environment.

While occupancy information is easily described by grid maps, semantic
information has on the other end a less established mapping tradition. In
semantic maps, labels are assigned to the environment in order to describe its
qualitative features. The labels could either be direct descriptions for objects
found in the environment, \eg{} ``chair'', ``plant'' and
``desk''~\cite{meger_curious_2008, costante_transfer_2013,
trevor_efficient_2013}, or more abstract descriptions of places, like
``office'', ``floor'' and ``kitchen''~\cite{pronobis_large-scale_2012,
mozos_categorization_2012}. For a more extensive survey over semantic mapping,
in~\cite{kostavelis_semantic_2015} multiple approaches are compared and
classified.

Models to describe an environment in multiple layers have been developed. The
Spatial Semantic Hierarchy \cite{kuipers2000spatial} represents an environment
on five levels that contain sensory, control, causal, topological and metric
data. While this model describes a multi-layered map, the layers are fixed and
related to robot navigation. Our goal is to develop a generalized framework able
to incorporate arbitrary information on a flexible amount of layers.
Specifically, an implementation for a polygonal semantic layer is provided,
which is not included in the SSH model.

Autonomous exploration for occupancy maps is a well-researched topic. A commonly
used approach is frontier exploration~\cite{yamauchi_frontier-based_1997}, which
aims to look for frontiers of the known maps to explore. Frontier cells are
defined as free cells that are adjacent to unknown cells. Newer research focuses
on exploration of 3D environments~\cite{shen_autonomous_2012} and on utilizing
learning methods to improve exploration behavior~\cite{zhu_deep_2018}. For
semantic maps, however, only few papers focus on autonomous exploration. Jebari
\etal{}~\cite{jebari_multi-sensor_2011} present an algorithm to autonomously
explore an unknown environment while simultaneously building occupancy and
semantic maps. They were however only interested in recognizing the positions of
a few known objects and not their spatial extent, which is a problem we address
in this work.
\section{Problem formulation}
\label{sec:problem}

A map is an entity that records the spatial location of objects in the
environment. Formally, let a map $\mathcal{M} = (P, O, \objs{}, c, s)$ be
defined by:
\begin{itemize}
	\item a position definition $P$, which is used to refer to the spatial
	location of objects;
	\item an object definition $O$, which usually includes a value $v \in V$,
	where $V$ is a set of possible values, and a position $p \in P$;
	\item a set of object entities $\objs{} \subseteq O$, following the object
	definition $O$;
	\item a ``content'' function $c: P \to O$ that, given a position, returns a
	set of objects at that position;
	\item a ``search'' function $s: V \to P$ that, given a value, returns the
	positions of objects having that value. 
\end{itemize}

To provide an example, an occupancy map $\mathcal{M}_\omega$ is most commonly
represented as a grid map. Therefore, each position $p_\omega \in P_\omega$ can
be defined as a row and column pair; formally, $P_\omega = \{(r, c) \mid r,c \in
\mathbb{N}\}$. Each position refers to a cell $o_\omega \in \objs{}_\omega$,
which has object representation $O_\omega = \{(v_\omega, p_\omega)\}$, where
$v_\omega \in [0,1]$ represents the probability of occupancy of the cell. From
this formulation, it can be noted that in grid maps there is no trivial way to
obtain the extent of an object for objects spanning several cells since each
cell is considered independently. On the other hand, semantic maps should store
labeled objects with their extent. Therefore, the grid representation is not
suitable for these maps, while polygonal maps are to be preferred. So, for
semantic maps, we define an object $o_\sigma \in \objs{}_\sigma$ by its label
and occupied area, \ie{} $O_\sigma = \{(v_\sigma, A_\sigma)\}$ with $v_\sigma
\in V_\sigma$ and $A_\sigma \subseteq P_\sigma$, where $V_\sigma$ is the set of
all class labels. The position definition in this case refers to coordinates in
a frame of reference, therefore $P_\sigma = \{(x, y) \mid x,y \in \mathbb{R}\}$.

Since the representations of the maps differ, an application needing to treat
information coming from multiple maps would need specific functions to handle
the different types of representations separately. Ideally, the application
should only communicate with a ``master'' map containing all knowledge about the
environment arising from different maps. In this work, we propose a framework
that provides a unified access method to maps of different types.

\section{Method}
\label{sec:method}

\subsection{\hmap{} framework structure}

Let us define a \hmap{} as $\mathbf{M}_H = (P_H, V_H, L)$, where $P_H$ and $V_H$
are position and value definitions used by the framework, and $L =
\{(\mathcal{M}_i, t_i, u_i)\}_{i=1}^N$ is a set containing the $N$ maps
$\mathcal{M}_i$ together with two interface functions $t_i$ and $u_i$. A
function $t_i: P_H \to P_i$ is used by the framework to convert from its own
position definition to the $i$-th layer's one, while a function $u_i : V_H \to
V_i$ converts from the framework value definition to the $i$-th layer's one.

The $t_i$ and $u_i$ functions are used by the framework's ``content'' and
``search'' functions, in the following way: the ``content'' function $\hat{c}$
allows to access the content of a set of layers $I = \{i_1, i_2, \dots, i_m\}$
at position $p_H \in P_H$, and is defined as:
\begin{align}\label{eq:contentH}
	\hat{c}(p_H, I) = \{c_i(t_i(p_H))\}_{i \in I}\enspace;
\end{align}
the ``search'' function $\hat{s}$ instead allows to look for the position of
objects having a certain value $v_H \in V_H$ in a set of layers $I$, and is
defined as:
\begin{align}\label{eq:searchH}
	\hat{s}(v_H, I) = \{t^{-1}_i(s_i(u_i(v_H)))\}_{i \in I}\enspace,
\end{align}
where $t^{-1}_i$ is the inverse function of $t_i$.

\begin{figure}
	\centering
	\includegraphics[width=\linewidth]{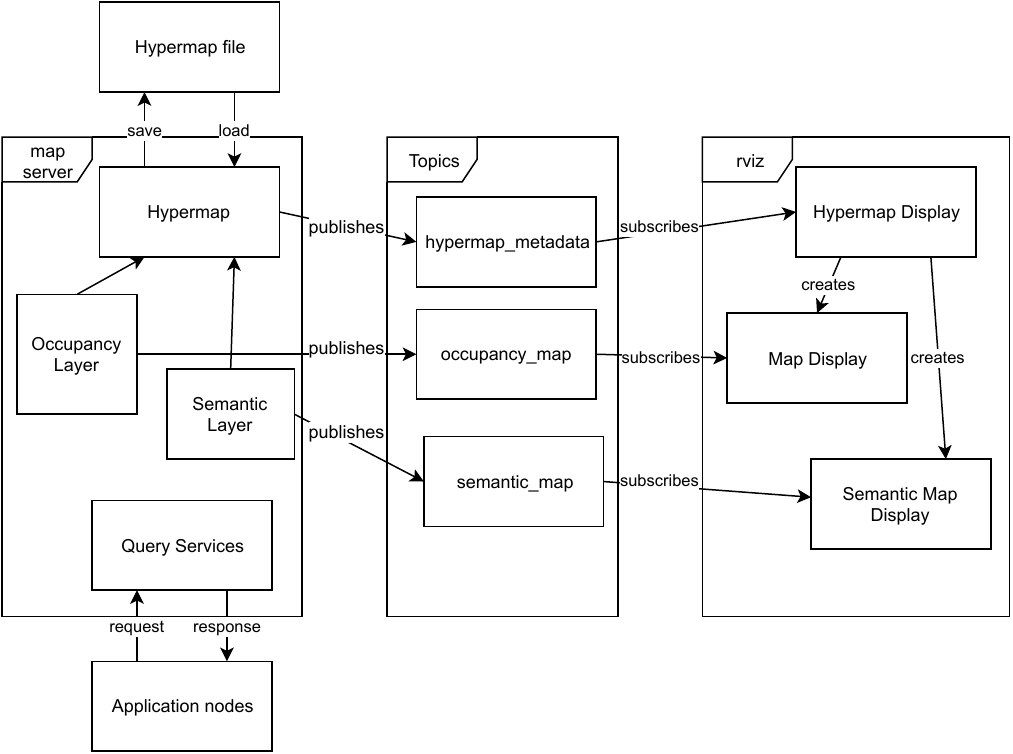}
	\caption{\label{fig:framework_overview}Framework implementation overview}
	\vspacefig{}
\end{figure}

\Cref{fig:framework_overview} shows an overview of the structure of the
framework implementation as a ROS node. The map server node stores the \hmap{}
with its layers. It can save and load \hmap{} files and provides services, which
allow other nodes to query the map for information. Furthermore, each layer
publishes its content to a topic and the metadata of the \hmap{} is published.
The \hmap{} display provided by the rviz plugin can subscribe to the metadata
topic. It automatically creates the necessary displays for the layers, which
subscribe to the respective topics.

For spatial queries, the global position definition $P_H = \{(x, y) \mid x,y \in
\mathbb{R}\}$ is used. In addition to point queries, area queries are possible.
The area can be specified by a list of points representing a simple polygon. The
global value type $V_H$ is a string. If a service for a spatial query is called,
the framework utilizes the content function to retrieve values from the layers,
while the search function is used if a value service is queried.

The occupancy and the semantic layers are represented as a grid map
$\mathcal{M}_\omega$ and a polygonal map $\mathcal{M}_\sigma$ respectively,
following the formalism presented in \Cref{sec:problem}. All conversion
functions are handled by the layer implementation. 

\subsection{Semantic mapping on polygonal maps}

While incremental mapping for grid maps is a well-studied problem, mapping on
polygonal maps has been studied less. As described in \Cref{sec:problem}, a
semantic object $o_\sigma$ consists of a label $v_\sigma$ and an area
$A_\sigma$. The area is defined by a list of vertices forming a polygon. During
the mapping process, new sensor information has to be integrated so that the
area of this polygon can be incrementally estimated. This poses three main
challenges:
\begin{enumerate*}
	\item after each new reading, the labeled objects in the scene have to be
	put on the map, with their coverage area estimated;
	\item when a part of the environment is observed again after some time, new
	readings should be integrated with the estimate of the area of already known
	objects;
	\item confidence over the existence of objects should be updated whenever an
	already mapped area is observed, and eventually objects with low confidence
	should be removed from the map.
\end{enumerate*}

\begin{figure}
	\centering
	\includegraphics[width=\linewidth]{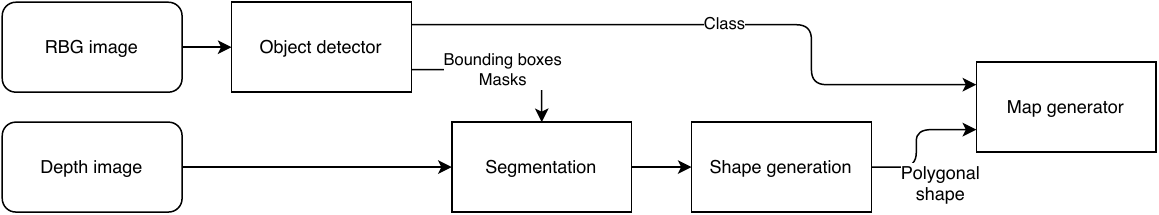}
	\caption{\label{fig:semantic_mapping}Semantic mapping}
	\vspacefig{}
\end{figure}

\begin{figure}
	\centering
	\begin{subfigure}[b]{.49\linewidth}
		\centering
		\includegraphics[width=\linewidth]{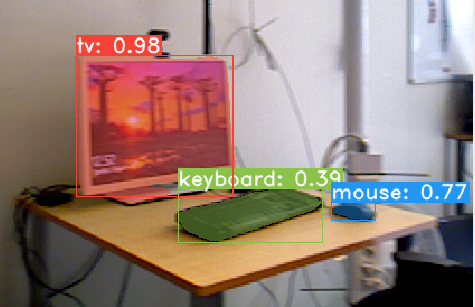}
		\caption{\label{fig:detector}Object detection}
	\end{subfigure}
	\begin{subfigure}[b]{.49\linewidth}
		\centering
		\includegraphics[width=\linewidth]
		{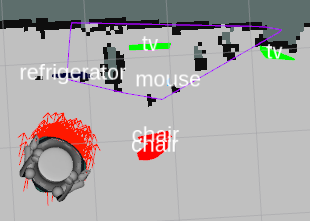}
		\caption{\label{fig:visibility}Visibility area}
	\end{subfigure}
	\caption{\label{fig:semantic}Examples of the semantic segmentation stages}
	\vspacefig{}
\end{figure}

The process of collecting evidence for the semantic mapping can be divided into
three phases: object detection, area generation, and map building.
\Cref{fig:semantic_mapping} gives an overview over the information flow from the
camera to the map generator. An RGB-D camera is used as sensor for the mapping.
The object detection is performed on the RGB image by an off-the-shelf deep
learning algorithm. For each detection, the algorithm provides a mask around the
object, as shown in \Cref{fig:detector}. It can be noticed that these masks can
sometimes include parts of the background. For this reason, the detected pixels
are transferred to corresponding points in a \pc{} generated from the depth
image of the RGB-D camera, and the background is removed by using a segmentation
algorithm.

Then, to determine the area of the object on the map, the \pc{} cluster
belonging to the object is projected on the $x$-$y$ plane of the map frame. The
transformation from the camera to the map frame has to be known, so a
localization on the existing occupancy layer is performed. From the projected
cloud, the convex hull is computed as object area. The area is passed on to the
map generator.

The generator looks for similar areas of the same class on the map. The
similarity is determined by computing the Jaccard index of the new areas with
any overlapping area on the map. The Jaccard index between area $A$ and $B$ is
computed as
\begin{align}\label{eq:jaccard}
	J(A,B) = \frac{|A \cap B|}{|A \cup B|}\enspace.
\end{align}
If the index exceeds a set threshold, the areas are assumed to belong to the
same object.

If an object is identified as a possible fit for the new area, the area is added
to the area list of the object and the existence probability of the object is
increased. Otherwise, a new object is created. If the existence probability
exceeds a set threshold, the object is considered part of the map.

To determine the area to display for the object, the average centroid of all
collected areas is computed. The area whose centroid is closest to the average
is chosen as the best fitting area and is displayed on the map.

To be able to remove falsely detected objects, the existence probability has to
be reduced in case an object is not detected. For this purpose, the visibility
area of the camera is used: for each object in the map within the visibility
area of a sensor measurement which has not received new evidence, the existence
probability is reduced. The visibility area is determined by projecting the
complete camera \pc{} to the $x$-$y$ plane of the map frame and then computing
the convex hull. An example of this visibility area is shown in purple in
\Cref{fig:visibility}.

\subsection{Exploration}

The autonomous exploration for semantic maps poses additional challenges
compared to the exploration for occupancy maps. Occupancy maps store which cells
are unknown. Therefore, exploration algorithms can determine from the map which
areas have yet to be explored. A polygonal semantic map, on the other hand, can
be sparsely populated with objects, which makes it impossible to differentiate
between unexplored and empty areas.

To keep track of which area has been explored, a new map type is introduced. An
exploration map $\mathcal{M}_\epsilon$ describes which parts of the environment
have already been explored. Since a known grid-based occupancy map is used for
navigation, it is beneficial to use a grid representation for the exploration
map. Therefore, $P_\epsilon = P_\omega = \{(r, c) \mid r,c \in \mathbb{N}\}$.
However, the value type differs from occupancy maps: instead of describing the
occupancy probability, $v_\epsilon \in \{0,1\}$ is 1 for explored cells and 0
for unexplored ones.

During the exploration, the values of the exploration map have to be updated. As
discussed in the previous segment, the current visibility area is determined
during the mapping process. However, this area uses the representation of the
semantic layer, that is, $A_\sigma \subseteq P_\sigma$. If the maps were used
separately, the conversion of the visibility area to the explored cells would
have to be performed on the application side. With the framework, the cells can
simply be accessed through the provided interface.

The mapping process continuously uses the currently observed area to update the
exploration map. An exploration application can access the values on this map to
determine which areas to explore next. Once all cells within a set area are
discovered, the exploration can be stopped.

\section{Experiments}
\label{sec:exp}

For the experiments, we first evaluated the capabilities of the semantic mapping
algorithm by generating maps through manually recorded data. We used a simulated
office created by Rasouli \etal{}~\cite{rasouli_effect_2017} as well as two real
environments, the robot laboratory and a floor. Furthermore, we wanted to
demonstrate the utility of using multiple layers of the framework. Therefore, we
show how the framework can be used to autonomously explore a limited area of an
existing occupancy map and populate it with semantic objects on a new layer.

\subsection{Platforms}

For the simulation, a simulated turtlebot was used. The real experiments were
performed on a \cob{} mobile platform. The robot is equipped with three laser
scanners to provide omnidirectional obstacle detection. The base can be moved in
any direction. Spherical joints also allow for 360\degree{} rotations without
moving the robot. Additionally, the robot contains multiple RGB-D cameras. For
the conducted mapping experiments, the camera placed on the sensor ring below
the head of the robot was used, as it provided the most complete view of the
environment in front of the robot.

The semantic mapping requires a localization on an existing occupancy map. The
occupancy maps of the environments were generated before the experiments from
odometry and laser scan data using off-the-shelf Gmapping
\cite{grisetti_improved_2007}. 

For object detection, a ROS wrapper for Yolact~\cite{yolact-iccv2019} was
implemented. Yolact offers real-time instance segmentation for images. No
training of the network was performed. Instead, the pre-trained base model
trained on the COCO dataset~\cite{lin_microsoft_2014} was used.

For the segmentation of the object \pc{}s to remove background points, PCL's
region growing algorithm was used. The resulting segment with the most points is
assumed to belong to the detected object.

The minimal Jaccard index for areas to be considered evidence for the same
object was set to 0.2. The value was chosen low as often, only parts of an
object are contained in the detected \pc{}\footnote{All the code, together with
the simulated environment used for experiments, is available here:
https://github.com/Eruvae/hypermap}.

\subsection{Environments}

\begin{figure}
	\centering
	\begin{subfigure}[b]{.74\linewidth}
		\centering
		\includegraphics[width=\linewidth]{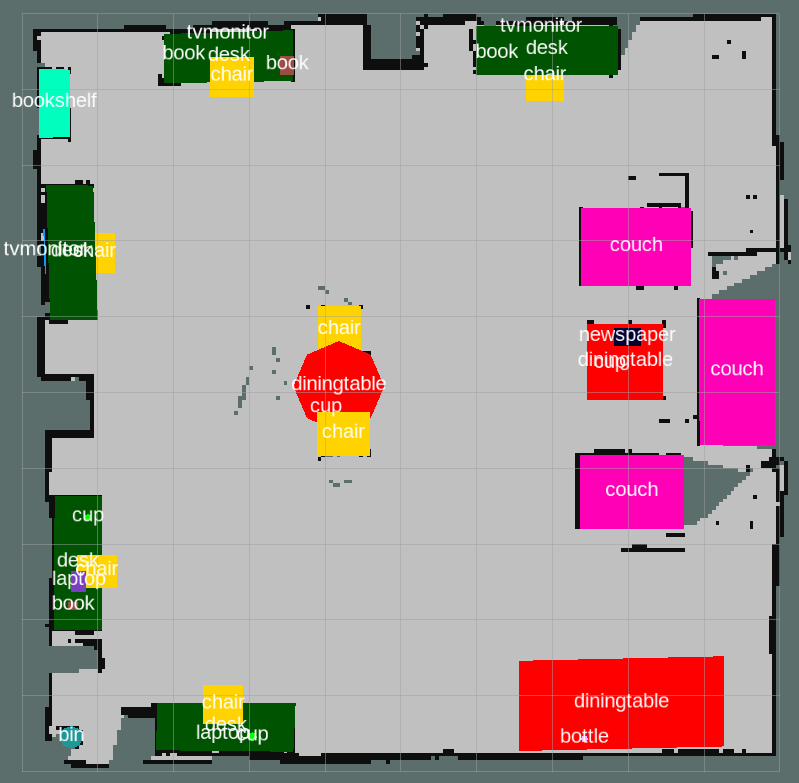}
		\caption{\label{fig:officegt}Ground truth}
	\end{subfigure}

	\begin{subfigure}[b]{.74\linewidth}
		\centering
		\includegraphics[width=\linewidth]{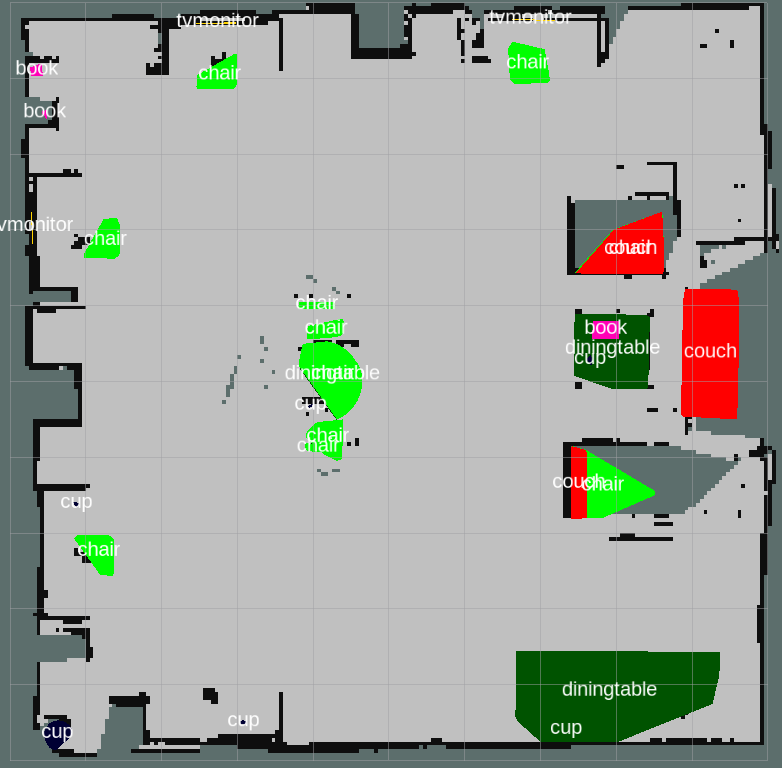}
		\caption{\label{fig:office_gen}Semantic segmentation}
	\end{subfigure}
	\caption{\label{fig:office}Simulated \eoffice{} environment}
	\vspacefig{}
\end{figure}

The simulated environment is an office with multiple desks and chairs. Monitors,
cups, and books are placed on the desks. Additionally, there are multiple dining
tables and a few couches. \Cref{fig:officegt} shows the manually labeled ground
truth map of the office. The duration of the mapping sequence was 134 s, during
which the robot moved around the whole office. We will refer to this environment
as \eoffice{}.

\begin{figure*}
	\centering
	\begin{subfigure}[b]{.45\linewidth}
		\centering
		\includegraphics[width=\linewidth]{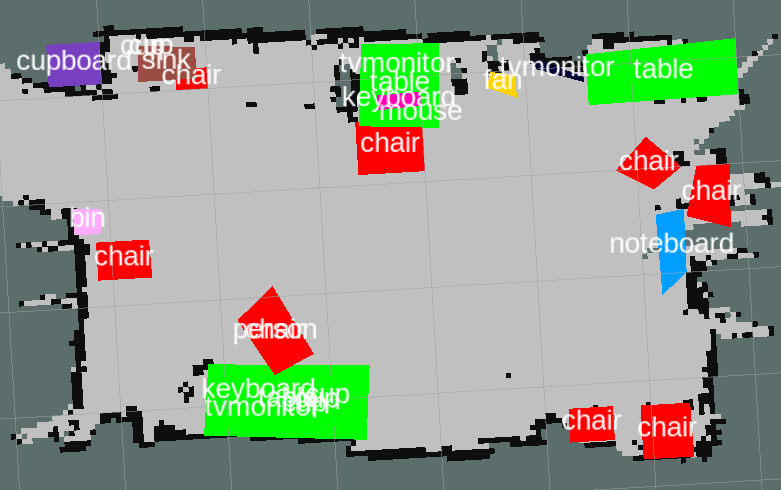}
		\caption{\label{fig:labgt}Ground truth}
	\end{subfigure}
	\begin{subfigure}[b]{.45\linewidth}
		\centering
		\includegraphics[width=\linewidth]{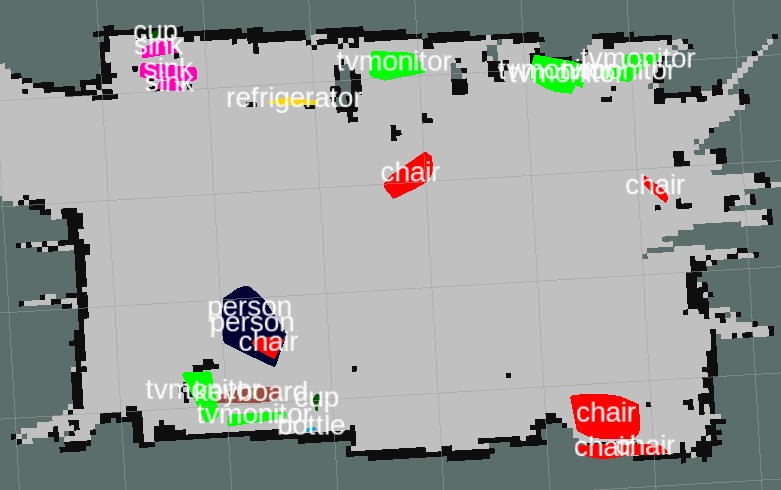}
		\caption{\label{fig:lab_gen}Semantic segmentation}
	\end{subfigure}
	\caption{\label{fig:lab}Real \elab{} environment}
	\vspacefig{}
\end{figure*}

The first real environment is a laboratory. \Cref{fig:labgt} shows the manually
labeled ground truth map. It contains multiple chairs, a sink, a cupboard and
tables with monitors, keyboards, mouses and cups. In the recorded sequence, the
robot moved for 143 seconds through the laboratory, moving from its home
position to the door and back while rotating a few times. We will refer to this
environment as \elab{}.

\begin{figure*}
	\centering
	\begin{subfigure}[b]{.49\linewidth}
		\centering
		\includegraphics[width=\linewidth]{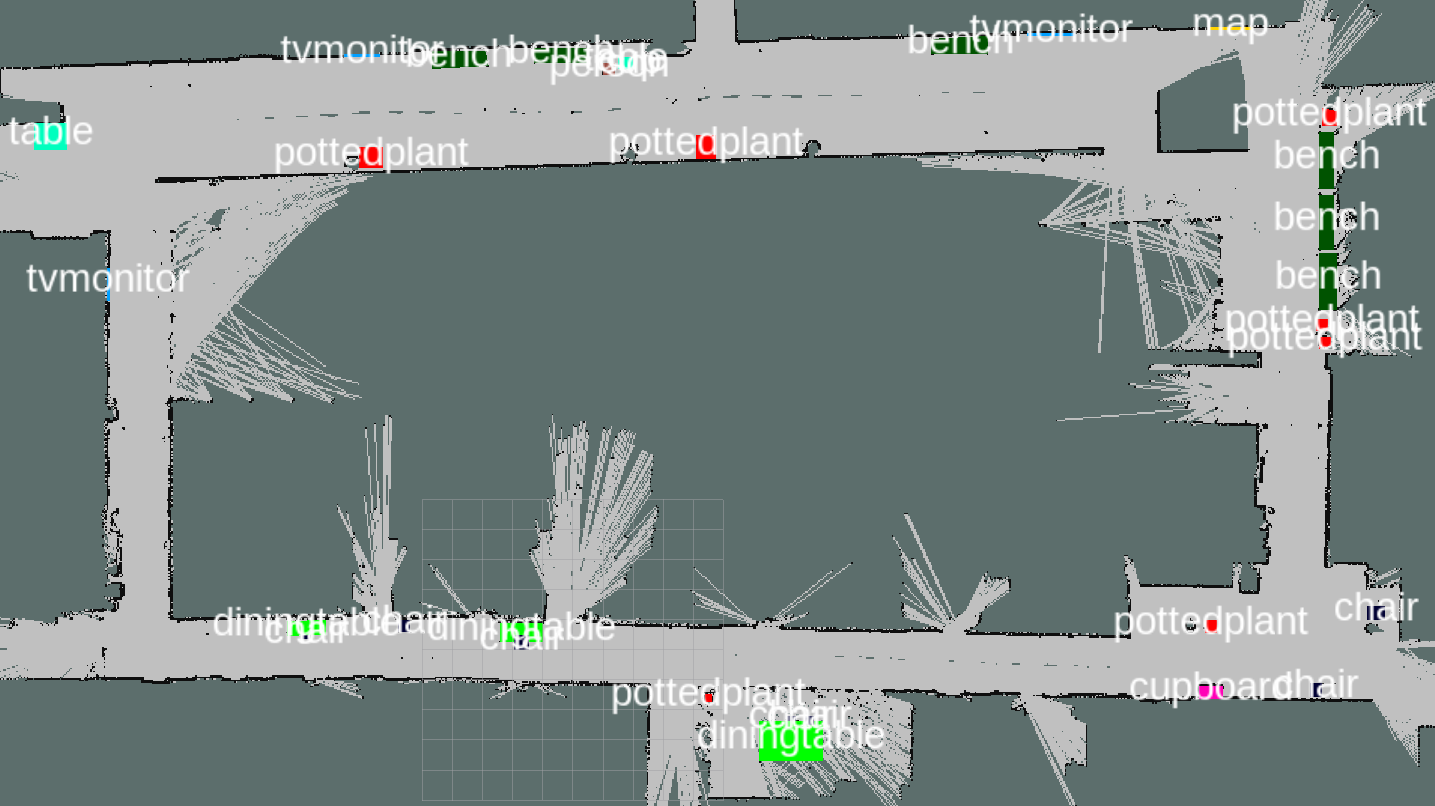}
		\caption{\label{fig:floorgt}Ground truth}
	\end{subfigure}
	\begin{subfigure}[b]{.49\linewidth}
		\centering
		\includegraphics[width=\linewidth]{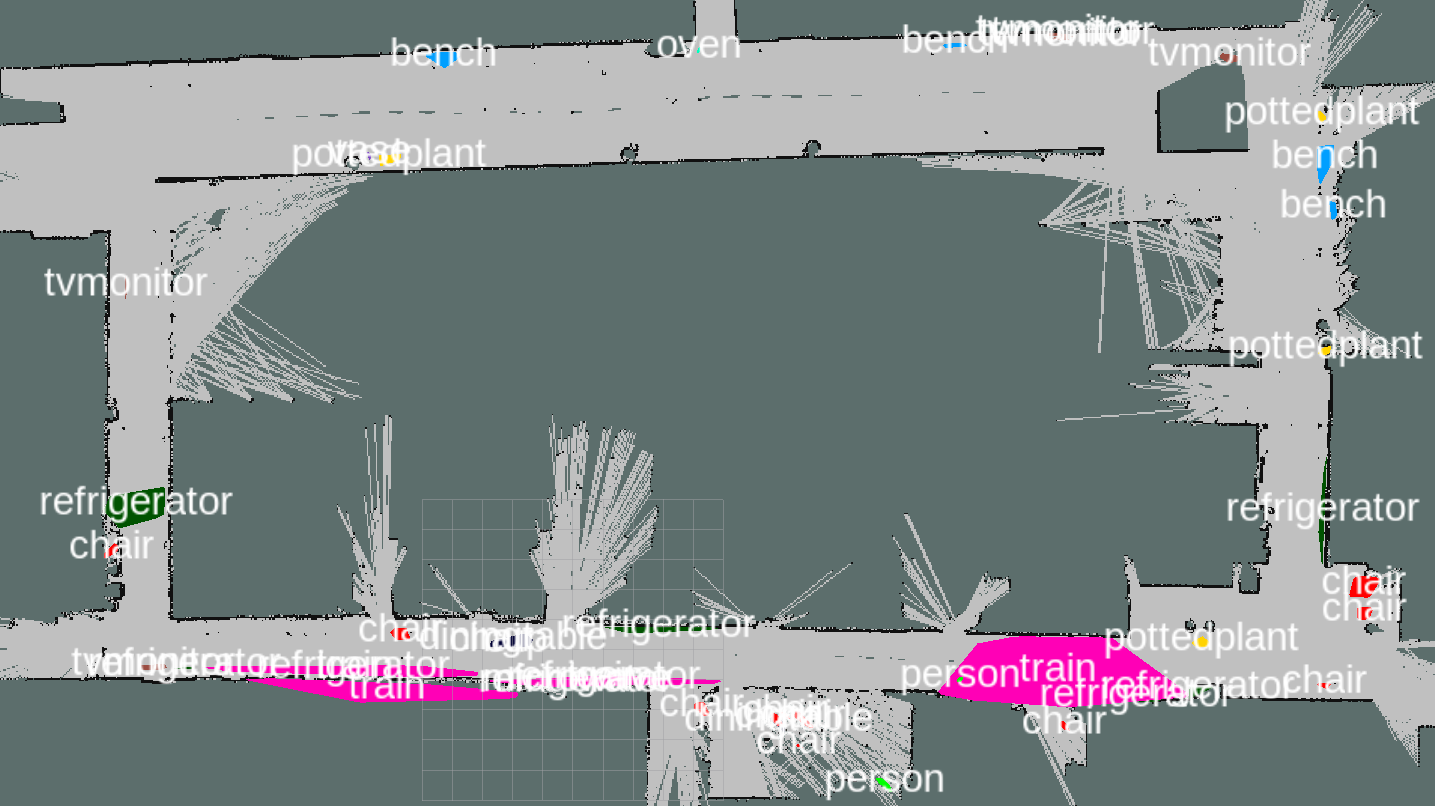}
		\caption{\label{fig:floor_gen}Semantic segmentation}
	\end{subfigure}
	\caption{\label{fig:floor}Real \efloor{} environment}
	\vspacefig{}
\end{figure*}

The second real environment is a floor of a university building.
\Cref{fig:floorgt} shows the corresponding ground truth map. The most common
types of objects are benches, chairs and potted plants, but the environment also
contains some tables and monitors. The recorded sequence was 7 minutes long,
during which the robot moved one time around the floor. We will refer to this
environment as \efloor{}.

The ground truth maps contain some objects that the neural network was not
trained to recognize, such as desks and cupboards. These objects were included
to provide a more complete overview of the environment but were ignored in the
detection analysis.

\subsection{Results}

The generated maps were compared to the ground truth maps. An object was counted
as detected if an object of that class was found within 1 m of the ground truth
object. For each successful detection, the quality was analyzed using two
parameters: the Jaccard index is computed as a measure of shape similarity, and
the centroid distance as a measure of localization accuracy.

\begin{table}
	\caption{\label{tab:office_gen}Detection results: \eoffice{}}
	\centering
	\begin{tabular}{lllll}
		\toprule
		& Ground truth & Detected & Jaccard & Centroid dist. \\
		\midrule
		\emph{diningtable} & 3 & 3 & 0.76 & 0.034 \\
		\emph{cup} & 4 & 4 & 0.41 & 0.087 \\
		\emph{couch} & 3 & 3 & 0.41 & 0.32 \\
		\emph{chair} & 7 & 7 & 0.53 & 0.19 \\
		\emph{tvmonitor} & 3 & 3 & 0.54 & 0.08 \\
		\emph{book} & 4 & 0 & - & - \\
		\emph{laptop} & 2 & 0 & - & - \\
		\emph{bin} & 1 & 0 & - & - \\
		\emph{bottle} & 1 & 0 & - & - \\
		\midrule
		Total & 28 & 20 & 0.52 & 0.15 \\
		\bottomrule
	\end{tabular}
	\vspacefig{}
\end{table}

\Cref{fig:office_gen} shows the map generated in the \eoffice{}. The detection
results are presented in \Cref{tab:office_gen}. 20 out of 28 detectable objects
were found, with an average Jaccard index of 0.52 and a centroid distance of
0.15 m. The map also contains some false detections. Two armchairs (labeled as
couch in the ground truth) were simultaneously recognized as chair and couch.
The central dining table was recognized as both a chair and a dining table. The
bin at the bottom left corner as well as the bottle on the bottom right dining
table were labeled as cups. While none of the books placed on the desks were
recognized, two books were found in the bookshelf on the top left corner.

\begin{table}
	\caption{\label{tab:lab_gen}Detection results: \elab{}}
	\centering
	\begin{tabular}{lllll}
		\toprule
		& Ground truth & Detected & Jaccard & Centroid dist. \\
		\midrule
		\emph{chair} & 8 & 5 & 0.27 & 0.23 \\
		\emph{tvmonitor} & 3 & 3 & 0.37 & 0.055 \\
		\emph{keyboard} & 2 & 1 & 0.63 & 0.077 \\
		\emph{mouse} & 1 & 0 & - & - \\
		\emph{sink} & 1 & 1 & 0.43 & 0.086 \\
		\emph{cup} & 5 & 2 & 0.48 & 0.073 \\
		\emph{person} & 1 & 1 & 0.80 & 0.23 \\
		\emph{bin} & 1 & 0 & - & - \\
		\midrule
		Total & 22 & 13 & 0.40 & 0.14 \\
		\bottomrule
	\end{tabular}
	\vspacefig{}
\end{table}

\Cref{fig:lab_gen} shows the detections in the \elab{} and \Cref{tab:lab_gen}
the corresponding results. Here, 13 out of 22 objects were detected, with an
average Jaccard index of 0.4 and a centroid distance of 0.14 m. The map contains
a few doubled objects; the sink was placed 3 times in different positions, and a
few monitors and as well as the person is recognized twice. A reason for this
could be localization inaccuracies of the robot, which causes the same object to
be seen in slightly different places.

\begin{table}
	\caption{\label{tab:floor_gen}Detection results: \efloor{}}
	\centering
	\begin{tabular}{lllll}
		\toprule
		& Ground truth & Detected & Jaccard & Centroid dist. \\
		\midrule
		\emph{pottedplant} & 7 & 4 & 0.44 & 0.44 \\
		\emph{diningtable} & 3 & 2 & 0.27 & 0.57 \\
		\emph{chair} & 7 & 5 & 0.48 & 0.30 \\
		\emph{bench} & 6 & 4 & 0.27 & 0.31 \\
		\emph{tvmonitor} & 3 & 2 & 0.32 & 0.40 \\
		\emph{person} & 1 & 0 & - & - \\
		\emph{cup} & 1 & 0 & - & - \\
		\midrule
		Total & 28 & 17 & 0.38 & 0.38 \\
		\bottomrule
	\end{tabular}
	\vspacefig{}
\end{table}

\Cref{fig:floor_gen} shows the map of the \efloor{}, with \Cref{tab:floor_gen}
for the corresponding results. 17 out of 28 objects were detected, with an
average Jaccard index of 0.38 and a centroid distance of 0.38 m. This map
contains more false detections than the previous environments. Parts of the
floor were sometimes recognized as trains by the neural network. These kinds of
detection faults could be prevented by training the network specifically for an
environment.

\subsection{Application: Autonomous Semantic Exploration}

The exploration capabilities of the framework were tested in the simulated
\eoffice{}. For the experiments, the exploration area was specified using an
rviz plugin for publishing polygons. After the border is set, the robot starts
exploring the area. To determine the exploration goals, a node designed for
occupancy map exploration~\cite{Lauri2016planning} is utilized.

\begin{figure}
	\centering
	\includegraphics[width=.8\linewidth]{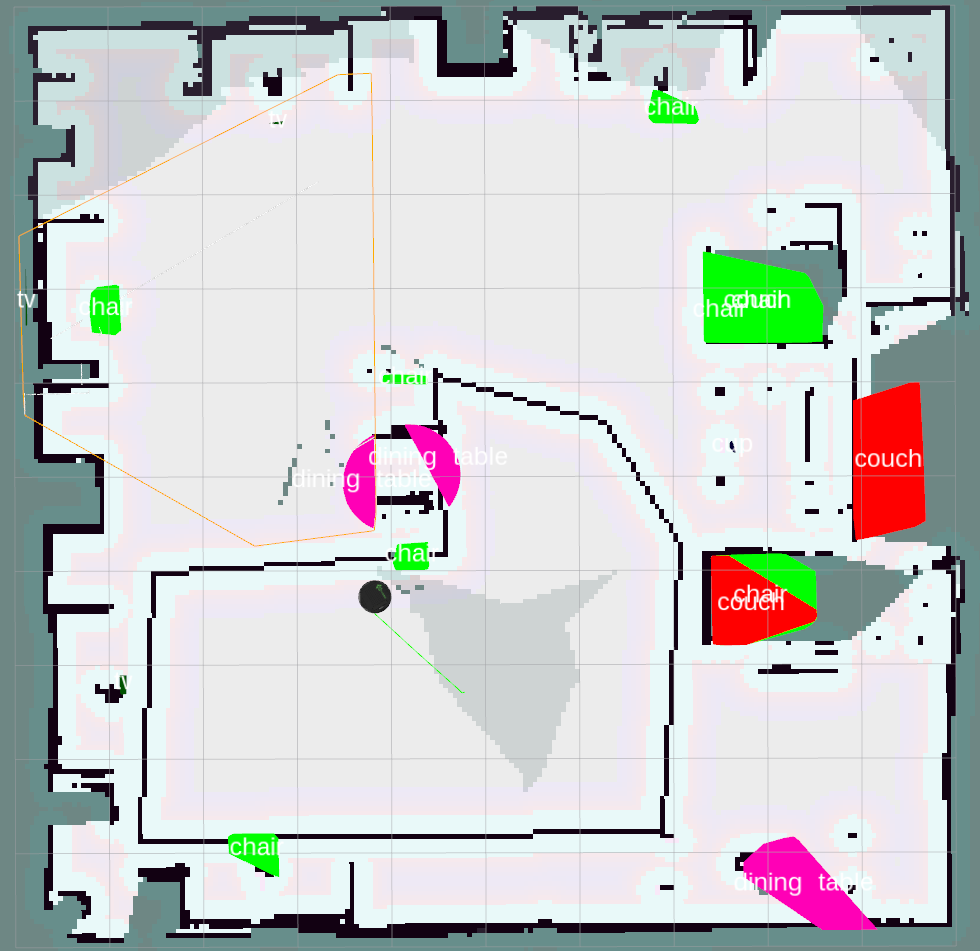}
	\caption{Explored semantic map}
	\label{fig:automap_res}
	\vspacefig{}
\end{figure}

\Cref{fig:automap_res} shows the resulting map for the described setup. The
bright cells mark the explored area, while the darker cells are unexplored. The
orange polygon shows the currently visible area. The exploration border is
marked by a polygon of occupied cells. While objects outside of the border are
still added to the map, the robot only moves within the marked area. Although
fewer objects were recognized than during the manually navigated tour around the
office, most objects close to the border were included in the map.

It is important to note that all interactions with the three maps required to
complete this task happened through the \hmap{} framework and the application
only required $P_H$ and $V_H$ to handle all the different underlying map
definitions.

\subsection{Discussion}

The experiments on the semantic mapping algorithm showed that it is capable of
estimating rough object positions and shapes. However, if the exact shape of
objects is needed, the method is not suitable yet, as sometimes, only parts of
the shapes are recognized. Furthermore, some objects are detected multiple times
in different positions.

One major problem is the dependency of the mapping on the localization through
the occupancy map. An algorithm that takes localization inaccuracies into
account and detects and removes duplicate objects could help improving the map
quality.

Furthermore, the approach could benefit from a neural network trained for office
environments. The COCO dataset is missing classes that are important to generate
complete semantic maps for offices, especially tables. On the other hand, it
contains classes that do not occur (e.g. trains), and therefore lead to false
detections.

For an accurate object shape representation, the shape combination would need to
be improved. Currently, one of the detected polygons is selected as the best
fitting shape. This can lead to incomplete shapes if an object was only
partially visible from most viewpoints. While merging overlapping shapes could
prevent that, it also leads to inflated shapes, especially in combination with
localization inaccuracies. A more intelligent approach would be necessary to
guarantee accurate and complete object shapes.

Regarding the autonomous exploration, we showed that the framework allows to
easily use existing nodes intended for occupancy exploration to semantically
explore parts of an existing map. While this demonstrates the benefits of a
multi-layered mapping framework, it does not exhaust its capabilities. Future
works could use the provided service to easily access other properties of the
environment.

\section{Conclusions}
\label{sec:concl}

In this work, we presented the \hmap{} framework, that allows to manage multiple
layers of different types of maps. We furthermore introduced a mapping algorithm
that creates polygonal semantic maps from RGB-D camera images. The semantic
mapping process has proven to be able to detect several different objects and
produce reasonable hypotheses for object placements and shapes. The framework
simplifies applications that depend on multiple map types, since they do not
need to handle different representations if they use the framework. As an
example, we demonstrated an autonomous mapping process that utilizes the
framework to provide and update an exploration layer while a semantic map is
being built. This process could update the exploration layer using polygonal
areas, even if internally a grid map was used.

For future work, the framework could be extended with new layer types for
different purposes. Furthermore, applications could be developed that use the
features provided by the framework to simplify tasks that require multiple types
of information about an environment.

\bibliographystyle{IEEEtran}
\bibliography{refs}

\end{document}